\pdfoutput=1
\documentclass[11pt]{article}
\usepackage[preprint]{acl}
\usepackage{times}
\usepackage{latexsym}

\usepackage[T1]{fontenc}
\usepackage{xspace}
\usepackage{booktabs}
\usepackage{multirow}
\usepackage{adjustbox}

\newcommand{\ours}[1]{\texttt{SimpleDoc}\xspace}
\usepackage{algorithm}
\usepackage[noend]{algpseudocode}  % <— suppresses EndIf/EndFor
\usepackage{amssymb}
\usepackage{amsmath}
\usepackage{enumitem} 

\usepackage[most]{tcolorbox}
\usepackage{xcolor}
\usepackage{listings}

\colorlet{select}{gray!20}
\definecolor{uclablue}{RGB}{0,69,126}
\definecolor{panton}{RGB}{216,27,96}
\definecolor{thupurple}{RGB}{102,0,153}
\definecolor{darkblue}{RGB}{0,0,139}
\definecolor{bigaired}{RGB}{205,92,92}

\colorlet{select}{gray!20}
\definecolor{uclablue}{RGB}{0,69,126}
\definecolor{panton}{RGB}{216,27,96}
\definecolor{thupurple}{RGB}{102,0,153}
\definecolor{darkblue}{RGB}{0,0,139}
\definecolor{bigaired}{RGB}{205,92,92}
\definecolor{sunny}{RGB}{253,184,19}  % new color for second box
\definecolor{brickred}{RGB}{203,65,84}  % frame color for second box

\newtcolorbox{llmprompt}[1][]{%
    enhanced,
    colback=select!50,
    colframe=uclablue!70,
    arc=1.5mm,
    boxrule=0.8pt,
    left=10pt, right=10pt, top=8pt, bottom=8pt,
    title={},
    listing options={%
        basicstyle=\ttfamily\small,
        breaklines=true,
        breakatwhitespace=false,
        showstringspaces=false,
        columns=flexible,
        keepspaces=true,
        upquote=true,
        frame=none,
        moredelim=[s][\color{panton}]{<}{>},
        moredelim=[s][\color{thupurple}]{<\\}{>},
        literate={User:}{{{\textcolor{darkblue}{\textbf{User:}}}}}{5}
                 {Assistant:}{{{\textcolor{bigaired}{\textbf{Assistant:}}}}}{10},
    },
    #1
}

\newtcolorbox{llmprompt2}[1][]{%
    enhanced,
    colback=sunny!30,
    colframe=brickred!70,
    arc=2mm,
    boxrule=1pt,
    left=10pt, right=10pt, top=8pt, bottom=8pt,
    title={},
    listing options={%
        basicstyle=\ttfamily\small,
        breaklines=true,
        breakatwhitespace=false,
        showstringspaces=false,
        columns=flexible,
        keepspaces=true,
        upquote=true,
        frame=none,
    },
    #1
}
\usepackage[utf8]{inputenc}

\usepackage{microtype}

\usepackage{inconsolata}

\usepackage{graphicx}

\title{SimpleDoc: Multi‑Modal Document Understanding with Dual‑Cue Page Retrieval and Iterative Refinement}
\author{
\textbf{Chelsi Jain}\textsuperscript{1,*}, \textbf{Yiran Wu}\textsuperscript{2,*}, \textbf{Yifan Zeng}\textsuperscript{1,3,*}, \textbf{Jiale Liu}\textsuperscript{2}, \\
\textbf{Shengyu Dai}\textsuperscript{4}, \textbf{Zhenwen Shao}\textsuperscript{4}, \textbf{Qingyun Wu}\textsuperscript{2,3}, \textbf{Huazheng Wang}\textsuperscript{1,3}\\
\textsuperscript{1}Oregon State University, \textsuperscript{2}Pennsylvania State University,\textsuperscript{3}AG2AI, Inc. ,\textsuperscript{4}Johnson \& Johnson \\
\texttt{\{jainc, zengyif, huazheng.wang\}@oregonstate.edu} \\
\texttt{\{yiran.wu, jiale.liu, qingyun.wu\}@psu.edu} \\
\texttt{\{SDai9, ZShao5\}@its.jnj.com} \\
}

\begin{document}

\maketitle
\def\thefootnote{*}\footnotetext{Equal Contribution.}\def\thefootnote{\arabic{footnote}}

\begin{abstract}

Document Visual Question Answering (DocVQA) is a practical yet challenging task, which is to ask questions based on documents while referring to multiple pages and different modalities of information, e.g., images and tables. To handle multi-modality, recent methods follow a similar Retrieval Augmented Generation (RAG) pipeline, but utilize Visual Language Models (VLMs) based embedding model to embed and retrieve relevant pages as images, and generate answers with VLMs that can accept an image as input. In this paper, we introduce \ours{}, a lightweight yet powerful retrieval‑augmented framework for DocVQA.  It boosts evidence page gathering by first retrieving candidates through embedding similarity and then filtering and re-ranking these candidates based on page summaries. A single VLM-based reasoner agent repeatedly invokes this dual-cue retriever, iteratively pulling fresh pages into a working memory until the question is confidently answered. \ours{} outperforms previous baselines by 3.2\% on average on 4 DocVQA datasets with much fewer pages retrieved.
Our code is available at \href{https://github.com/ag2ai/SimpleDoc}{https://github.com/ag2ai/SimpleDoc}.

% In this paper, we further enhance the retrieval of relevant pages by using an LLM to re-rank the retrieved pages.
% We build a reasoner agent that actively decides if all relevant information is retrieved, and will send query to ask for more information as needed. \todo{add experiment results} \todo{name and title}

% iteratively refine strategy, to retrieve more information if the VLM
% of what is missing based on 

% Answering questions from long documents

% (such as PDFs containing text, tables, and images)

% Most existing systems follow a one-step pipeline: they retrieve relevant document segments and generate answers in a single pass. However, real-world questions often require multi-step reasoning, referencing multiple document sections or sources, and integrating different modalities. 
% \yr{Our target task: Single Document with multi-modal information, Multiple Reference Pages, } 
\end{abstract}

\section{Introduction}
% \yr{Para 1: Intro to VQA} 
Documents are a fundamental form for the preservation and exchange of information, and an important source for humans to learn and acquire knowledge~\cite{gu2021unidoc, chia2024m, longdocurl}. Document question answering is a core task for automated understanding and retrieval of information~\cite{appalaraju2021docformer, van2023document}. Document Visual Question Answering (DocVQA) involves answering questions grounded in multi-modal documents containing text, tables, and images — common in formats like reports and manuals~\cite{suri2024visdom, mmlongbench}. There are three main challenges in this task: (1) \textit{multiple pages},  where a portion of a long document needs to be processed to answer the question, (2) \textit{multiple references}, where different pages need to be cross-referenced, and (3) \textit{multiple modalities}. %(1) Long documents retrieval: a document can contain many pages, which is a large chunk of information to go through to answer the question. Asking questions based on multi-documents would be even harder~\cite{m3docrag, saad2023pdftriage};  (2) Multiple pages might need to be accessed and referenced; (3) Different modalities of information need to be referenced. 

% \begin{figure}[t!]
%     \centering
%     \includegraphics[width=0.48\textwidth]{figures/first_figure.pdf}
%     % \caption{Overview of vanilla RAG and our method. \todo{@Chelsi: this is a prelimnary figure, you should still need modification. Note we don't need to plot MDocAgent now.}}
%     % \label{fig:rag_baseline}
% \end{figure}

\begin{figure}[t!]
    \centering
    \includegraphics[width=0.47\textwidth]{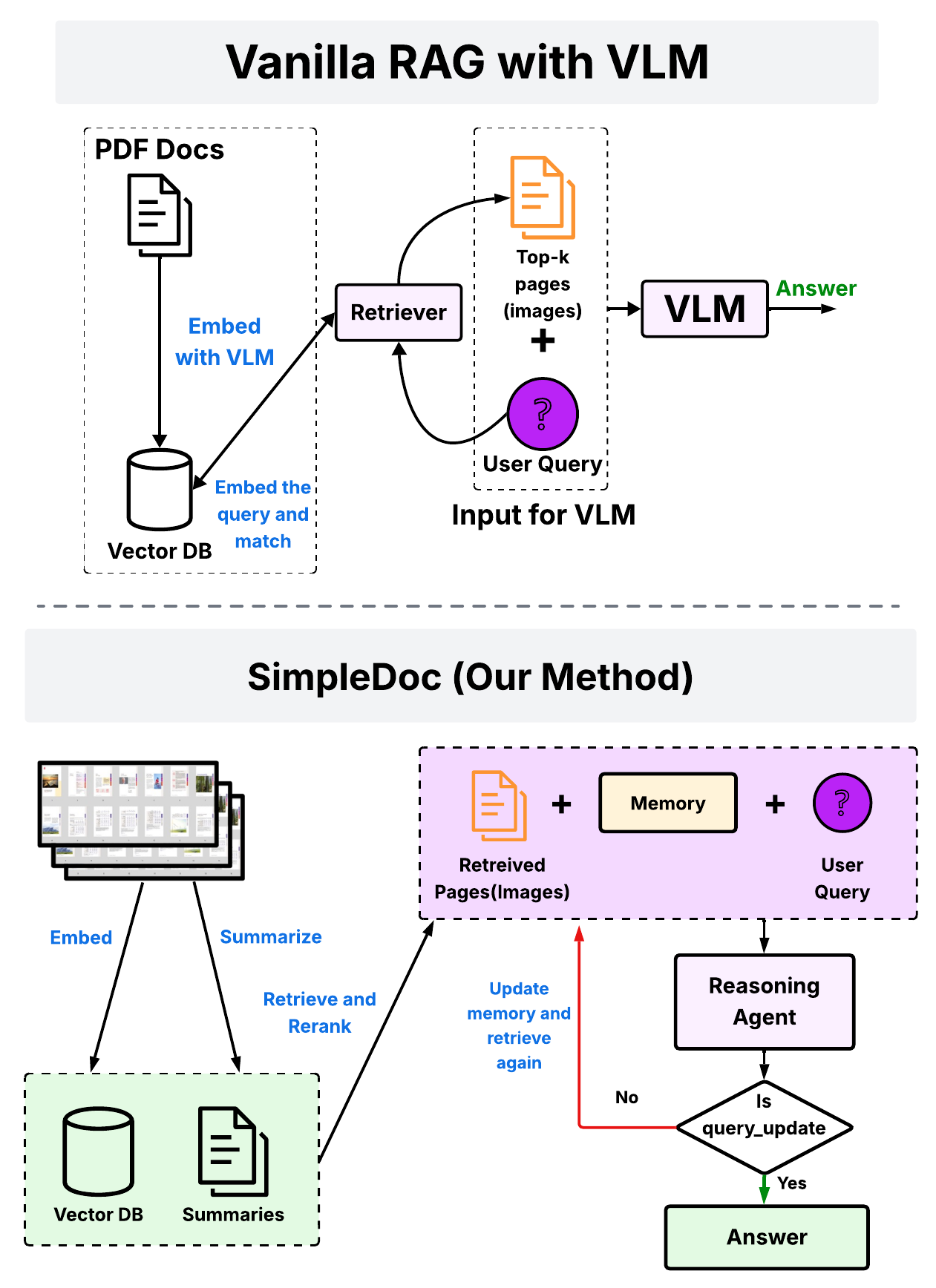}
    \caption{Illustration of the vanilla Retrieval-Augmented Generation (RAG) pipeline and the proposed \ours{} framework. \ours{} introduces a two-step page retrieval process that utilizes pre-processed embedding and summaries of each page. During generation, a reasoning agent reviews the retrieved pages and decide whether to give the answer, or produce a new query to retrieve more pages.}
    \label{fig:rag_baseline}
    \vspace{-0.4em}
\end{figure}

% \begin{figure}[t!]
%     \centering
%     \includegraphics[width=0.48\textwidth]{figures/figure1.pdf}
%     \caption{Overview of vanilla RAG and our method. \todo{@Chelsi: this is a prelimnary figure, you should still need modification. Note we don't need to plot MDocAgent now.}}
%     \label{fig:rag_baseline}
% \end{figure}

 % \todo{@Yifan explain LLM rank and cite, like \cite{chen2024mllm, zhuang2023open}}

Retrieval-augmented generation (RAG)~\cite{lewis2020retrieval} is an effective pipeline to overcome challenges (1) and (2), where relevant information is retrieved by a retrieval model and then fed to a generation model to output the answer. To handle different modalities, several methods have been proposed to pre-process documents by converting different modalities into texts~\cite{memon2020handwritten, pypdf2, pdfminer_six}. Recently, multi-modal retrieval models such as CoPali~\cite{colpali} are proposed to perform page-level retrieval by treating each page as image~\citep{yu2024visrag, xie2024wukong}. Building on this, M3DocRAG \cite{m3docrag} proposed a multi-modal RAG system that demonstrated strong performance in DocVQA tasks by combining image and text embeddings for document retrieval. Since multi-agent systems have emerged as an effective method to solve complex tasks and multi-step tasks~\cite{autogen, deepresearcher, wu2024stateflow}, MDocAgent \cite{han2025mdocagent} applied this concept to document QA by designing a multi-agent pipeline composed of dedicated text and image retrieval agents, a critical information extractor, and a final summary agent to collaboratively tackle multi-modal document understanding. Despite MDocAgent's effectiveness, we find it to be overcomplicated and might not utilize the full capacity of recent VLMs. 

\ours{} introduces a simple retrieval augmented framework that leverages modern VLMs without the overhead of complex multi‑agent designs.  The pipeline unfolds in two stages.  First, an offline document‑processing stage indexes every page twice: (i) as a dense visual embedding produced by a page‑level VLM such as ColPali, and (ii) as a concise, VLM‑generated semantic summary that captures the page’s most salient content.  Second, an online iterative QA stage employs a dual‑cue retriever that initially shortlists pages via embedding similarity and then asks an LLM, which operates solely over the summaries, to decide which of those pages are pertinent to the query and re-rank them by estimated relevance. This ordered subset is handed to a single reasoning agent. The agent reads only the newly selected pages along with a working memory, which preserves important information from previously examined pages, and judges whether the evidence now suffices to answer the question.  If it detects missing information, the agent emits a refined follow‑up query, prompting another retrieval round and merging the newly distilled notes into memory.  This lightweight loop of targeted retrieval and memory‑aided reasoning continues until an answer is produced or a preset iteration limit is reached, enabling \ours{} to flexibly trade retrieval depth for generation quality.

% \yr{Para 4: Experiments and results} 
We perform various experiments and analyses to gain an understanding of the VQA problem and to validate the effectiveness of our method.  We test on 4 different datasets and find that our method can improve over previous baselines by 3.2 absolute points, with only 3.5 pages retrieved for each question. While the setting of multi-modal, multi-page document-based QA seems new, we find it very much resembles `traditional' RAG tasks focusing on tasks like HotpotQA~\cite{yang2018hotpotqa} and 2WIKI~\cite{ho2020twiki}, which usually require retrieved fine-grained chunked texts from given documents. However, M3DocRAG and MDocAgent have had few discussions in this direction. Instead, we do a detailed analysis of these RAG methods and uncover two common strategies: query decomposition and relevant page review. We implement Plan$^*$ RAG and Chain-of-note as representations of the common strategies and compare them under the DocVQA setting. To summarize, our contributions are the following:
\vspace{-0.4em}
\begin{itemize}[leftmargin=*]
\setlength\itemsep{-0.1em}
    \item We propose \ours{}, a straightforward and effective framework for multi-modal document question-answering.
    % a new method to retrieve documents with VLMs, and iteratively retrieve more documents as needed.
\item We perform various experiments to test effectiveness of \ours{}, and analyze and compare with traditional RAG methods in which previous methods on DocVQA are missing.
% and how recent VLMs perform on this task.
\end{itemize}
 % We propose a new method improving both retrieval and generation: 1. For retrieval, we use VLM to summarize each page (preprocess), and fit all summaries to do a re-ranking. \yr{Maybe add the embedding retrieval part} 2. For generation, we use one VLM to generate the answer. At the same time, we ask it to determine if any documents are missing, and do another retrieval if needed.  

% Figure~\ref{fig:rag_baseline} illustrates this pipeline, where retrieved documents and images are passed to a VLM to generate answers.

\begin{figure*}[t!]
    \centering
    \vspace{-2mm}
    \includegraphics[width=0.95\linewidth]{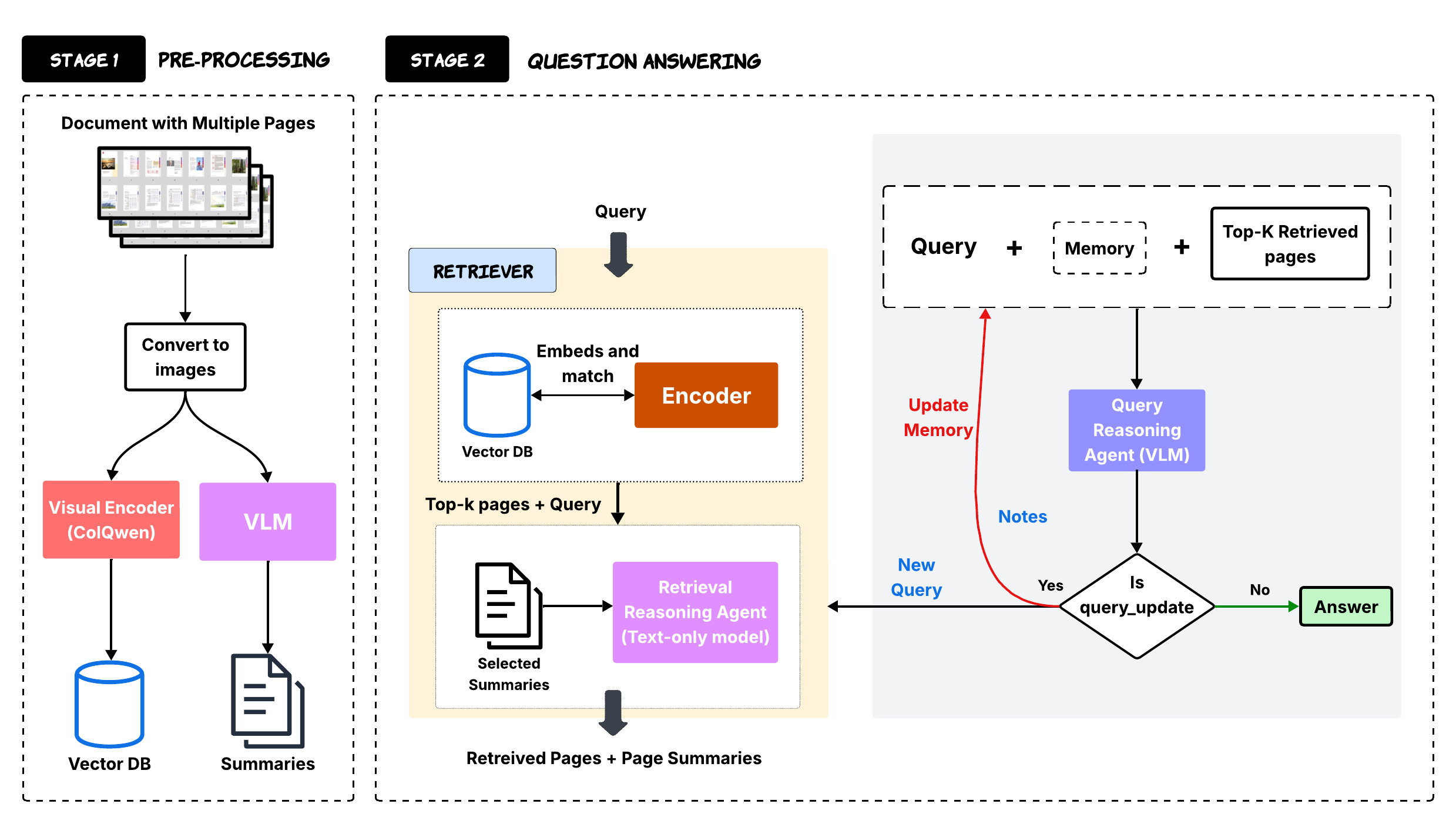}
    \caption{\ours{} consists of two stages: (1) offline extraction of visual embeddings and LLM-generated summaries for all document pages, and (2) an online reasoning loop that performs retrieval via embedding and summary-based re-ranking, followed by answer generation with a memory-guided VLM agent that iteratively refines its query if needed.
    }
    \label{fig:main_figure}
    \vspace{-4.5mm}
\end{figure*}

\vspace{-0.6em}
\section{Related Work}
\vspace{-0.3em}

\textbf{Document visual question answering.} focuses on answering questions grounded in visual and textual information contained within documents~\cite{ding2022v, tanaka2023slidevqa}. Early efforts primarily addressed single-page document images using OCR-based approaches and multi-modal language models (MLMs)~\cite{docvqa, infographicvqa, ocrvqa}. However, these methods often struggled with the long-context reasoning and complex layouts found in real-world documents. Recently, benchmarks like MP-DocVQA~\cite{mpdocvqa} and MMLongBench-Doc \cite{mmlongbench} focus on long multi-page and multi-modal document understanding, posting new challenges to the task~\cite{tanaka2023slidevqa}. However, recent advances in vision-language models (VLMs) has shown promise for multi-modal document understanding~\cite{liu2024improved, liu2023visual, chen2022murag, bai2025qwen2, pdfwukong, ma2024visa}. ColPali~\cite{colpali} introduces a new concept of treating document pages as images to produce multi-vector embeddings, where pages can be retrieved for each query. Other methods such as VisRAG~\cite{yu2024visrag} and VDocRAG~\cite{tanaka2025vdocrag} also convert pages as images to avoid missing information from parsing text and image separately from one page. From CoPali, M3DocRAG \cite{m3docrag} proposed a multi-modal RAG pipeline that retrieves relevant document pages across large document corpora and feeds them into a vision language model.
MDocAgent~\cite{han2025mdocagent} extended this by introducing specialized agents for handling cross-modal retrieval and reasoning over long documents.

% Compared with prior work, \ours{} uses a LLM to further filter and re-rank the retrieved pages using summaries of each pages to locate
% to retrieve more relevant pages.
% retrieving top-k relevant pages, summarizing their content with respect to the query, and optionally decomposing complex questions into sub-queries to predict intermediate answers, which are then aggregated into a final response. 

 % ma2024visa

% Other methods such as PDFWuKong \cite{pdfwukong} and  are also proposed to 

% \textbf{Multi-modal Learning and LVLMs for Document Understanding.} R Qwen2-VL \cite{qwen2vl}, and SlideVQA \cite{tanaka2023slidevqa} demonstrate strong performance on vision-language benchmarks. VDocRAG \cite{vdoc} introduced document understanding by representing entire documents as images to avoid text parsing errors. While these approaches offer powerful capabilities, they often lack structured document navigation and struggle with extremely long document contexts.
% \vspace{-0.1em}
\noindent\textbf{Retrieval augmented generation (RAG)}  
has become a powerful strategy for knowledge-intensive tasks by supplementing language models with external context, which consists of two core steps: retrieve and generate~\cite{jiang2023active, gao2023retrieval}. Many works have been proposed to improve RAG, such as training effective embedding models~\cite{karpukhin2020dense, colbert}, query rewrite and decomposition~\cite{ma2023query, peng2024large, rqrag, planrag, lee2024planrag, wang2024e5}, constructing different forms of databases (e.g., knowledge graphs)~\cite{gaur2022iseeq, edge2024local, liu2025hm}, improving quality of retrieved context~\cite{chainofnote, chen2024dense}, augmenting the RAG process~\cite{selfrag, trivedi2022interleaving, liu2024ra}, and many others~\cite{activeretrievalaugmentedgeneration}. Most of the RAG methods focus on knowledge and reasoning tasks that only require text-based retrieval (e.g., HotpotQA)~\cite{yang2018hotpotqa, geva2021strategyqa, trivedi2022musique, mallen2023popqa, ho2020twiki, kwiatkowski2019nq}. While we are targeting the Document Visual understanding task, we find that many core ideas might also be effective in DocVQA. Thus, we also implement and test two RAG methods: Chain-of-Notes~\cite{chainofnote}, which improves retrieval context for better generation, and Plan$^*$RAG~\cite{planrag}, which decomposes queries and augments the generation process for better retrieval, to help understand how previous methods can be used on DocVQA.

\vspace{-0.4em}
\section{Method}
\label{sec:method}
\vspace{-0.3em}
% For Document QAs, there are mainly two parts: 1. Retrieve evident pages based on the question. 2. Generate answers based on the retrieved questions.
% \begin{enumerate}
%     \item Evidence Page + Answering
%     \item Retrieved top k (MDoc's Retrieve method) + Answering
%     \item Retrieved top k + Reranking + Answering
%     \item Retrieved top k + Reranking + Answering - > If need more pages, generate sub queries and retrieve (iterative refine)
% \end{enumerate}

% Iteratively retriveal and 
% can efficiently handle long documents by using an iterative retrieval process that dynamically locates relevant information for answering user queries. 
Below we introduce \ours{}, an effective framework for DocVQA. \ours{} consists of two stages: an offline document processing phase followed by an online iterative retrieval-augmented question answering phase. Our framework features the following: 1. Enhanced page retrieval through a combination of vector and semantic representations. 2. Continuous refinement via iterative retrieval and memory update. Figure~\ref{fig:main_figure} illustrates the overall pipeline of our approach.
\vspace{-0.3em}
\subsection{Offline Document Processing}
\vspace{-0.3em}
The initial stage involves pre-processing and indexing each document to create a searchable representation. We treat each page as a unit, and use two VLMs to get both vector and semantic representations of each page.
For vector embedding, we employ VLM like CoPali~\cite{colpali} that are trained to generate embeddings for document pages. For semantic representation, we use a general VLM guided by a predefined prompt to produce a summary (typically 3-5 sentences) that includes the salient information of that page. These summaries are designed to highlight information that might be generally relevant for answering potential future questions without prior knowledge of any specific user query. 

Specifically, given a document $D$ consisting of $j$ pages $D = {p_1, p_2, ..., p_j}$, we use a vision embedding model to generate embedding vectors $E = \{e_1, e_2,\dots, e_j\}$ for each page, and use a VLM to generate $j$ summaries $S = \{s_1, s_2, ..., s_j\}$.
\vspace{-0.2em}
\subsection{Multi-modal Question Answering}
For retrieval, we use a VLM to retrieve pages through embedding similarity, and a VLM to look at the summaries and re-rank those retrieved pages. During the question answering phase, we build a reasoner agent that can automatically decide whether to retrieve more information and iteratively refine its own memory with newly retrieved pages. 

\vspace{-0.2em}
\paragraph{Page Retrieval}
Given a query $q$ and its document $D$, we first embed the given query and retrieve $k$ pages with the highest MaxSim score \cite{khattab2020colbert}. Then, we pass $q$ and $k$ summaries of the retrieved pages $S_k$ into an LLM (can be text-only) to select and rank the relevant pages. The model returns an ordered list of page indices $C={c_1,c_2,\dots,c_n}$ based on their perceived relevance to the query. Note that the number of relevant pages is automatically and dynamically chosen by the model. Since the re-rank is based on the retrieved pages from embedding, so $n<k$ pages are later sent to the reasoner agent, keeping the input size manageable. In this step, we also ask the LLM to generate an overall document-level summary $s_{\textsc{doc}}$ that contextualizes the entire document in relation to the current query, serving as the initial working memory of the reasoner agent.

\begin{algorithm}[ht!]
\caption{\ours{}}
\label{alg:iterative-qa}
\begin{algorithmic}[1]
  \Require query $q$, per–page embeddings $E$ and summaries $S$, cutoff $k$, max iterations $L$
  \Ensure answer $a$ or failure notice
  % page images $I$ and page texts $T$ from initial retrieval, global summary $s_{\textsc{doc}}$,
  \State $q_{\text{cur}} \gets q$
  \State $M \gets \varnothing$
  \For{$\ell \gets 1$ \textbf{to} $L$}
      \State $s_{\textsc{doc}}, C \gets RetrievePages(q_{\text{cur}}, E, S, k)$
      \State $I_C \gets \{\,i_c \mid c \in C\}$;\;
             $T_C \gets \{\,t_c \mid c \in C\}$
      \State $M \gets M \cup s_{\textsc{doc}}$
      \State $(\textit{is\_solved}, a, m', q')\gets$
      \State \;\;\;\;\;\;$\textsc{Reasoner}(q, I_C, T_C, M)$
      \If{$\textit{is\_solved}$}
         \State \Return $a$
      \Else
         \State $M \gets M \cup \{m'\}$
         \State $q_{\text{cur}} \gets q'$
      \EndIf 
      % \State $C \gets RetrievePages(q_{\text{cur}}, E, S, k)$
  \EndFor
  \State \Return \textsc{Fail}
\end{algorithmic}
\end{algorithm}

\vspace{-0.2em}
\paragraph{Generation} We treat the retrieved relevant pages as images, denoted as $I_C=\{i_{c_1},i_{c_2},\dots,i_{c_n}\}$. Those pages are also converted into text, denoted as $T_C=\{t_{c_1},t_{c_2},\dots,t_{c_n}\}$. We input will $I_C$, $T_C$, input query $q$ and a working memory $M$ (initialized to $s_{\textsc{doc}}$) into a reasoner agent (backed by a VLM), and ask it to determine if the question can be solved with the given context. 

The reasoner can produce one of three distinct response types:
\vspace{-0.4em}
\begin{itemize}[leftmargin=*]
\setlength\itemsep{-0.1em}
    \item \textbf{Answer}: If the provided pages contain sufficient information, the reasoner formulates a direct answer to the query.
    \item \textbf{Not Answerable}: If the question cannot be answered by the document.
    % the document does not contain the required information
    \item \textbf{Query Update}: If the reasoner believes the answer exists within the document but on pages not yet retrieved, it outputs a note of current pages $m'$ and generates a new query $q'$ that asks for missing  information.
\end{itemize}
\vspace{-0.8em}
\paragraph{Iterative Refinement} Self-reflection has been proven an effective method in LLMs~\cite{shinn2023reflexion, madaan2023self}. We employ a similar mechanism where the LLM can actively retrieve more pages as needed. If the reasoner agent decides that the question cannot be solved after the initial retrieval, we start an iterative process to continue retrieving new pages. As shown in Algorithm~\ref{alg:iterative-qa}, we maintain a memory module $M$ to preserve useful information from previous retrievals. When the reasoner agent outputs a query update, we retrieve new page numbers $C'$ based on the refined query $q'$, update the memory module $M$ with the notes $m'$, and call the reasoner again with the following inputs: $\{q, I_{C'}, T_{C'}, M\}$. The iterative process terminates when the reasoner produces an answer, determines the query is not answerable, or reaches a predefined maximum number of iterations $L$. If the maximum iterations are reached without resolution, the question is marked as "not answerable."

% based on the aggregated page summaries. This summary also serve as the high-level plan for solving the user query.
% Our retrieval consists of 2 parts: 1. treat PDF pages as images and use Copali as embedding to retrieve. 2. LLM Re-ranking: we first process and summarize the pages, and ask LLM to re-rank retrieved pages. LLMs can reject pages that are not related at all. 
\vspace{-0.6em}
\section{Experiments}
\vspace{-0.3em}

Our experiment is organized as follows: In Section~\ref{sec:mainresult}, we present the main results of our method and baselines on 4 different datasets. In Section~\ref{sec:diff_model}, we further experiment on MMLongBench using different models. In Section~\ref{sec:other_rag}, we adopt and implement two other RAG methods that were originally proposed for knowledge Question Answering, Finally inn Section~\ref{sec:additional_result}, we test variations of \ours{} and further analyze our method.

% In Section~\ref{sec:pilot}, we conduct a pilot experiment to understand how VLM performs with the golden evidence pages, and how different forms of evidences affect the performances. 

\begin{table*}[t!]
    \centering
    \begin{adjustbox}{width=1\textwidth}
    \begin{tabular}{lc|ccccc}
        \toprule
        \textbf{Method} & Pg.~Ret. & MMLongBench & LongDocUrl & PaperTab & FetaTab & \textbf{Avg. Acc} \\
        \midrule
        \multicolumn{7}{c}{\textit{LVMs}} \\
        % \midrule
        Qwen2.5-VL-32B-Instruct  & -- & 22.18  & 19.78  & 7.12 & 16.14  &  16.31  \\
        Qwen2.5-VL-32B-Instruct + \textcolor{blue}{Ground-Truth pages} & -- & 67.94 & 30.80 & - &  - & - \\
        \midrule
        \multicolumn{7}{c}{\textit{RAG methods (top 2)}} \\
        % \midrule
        M3DocRAG (Qwen2.5-VL-32B) & 2 & 41.8 & 50.7 & 50.1 & 75.2 & 54.4 \\
        MDocAgent (Qwen3-30B + Qwen2.5-VL-32B) & 4 & 50.6 & 56.8 & 50.9 & 80.3 & 59.6 \\
        \midrule
        \multicolumn{7}{c}{\textit{RAG methods (top 6)}} \\
        % \midrule
        M3DocRAG (Qwen2.5-VL-32B) & 6 & 41.8 & 53.1 & 60.1 & 79.8 & 58.7 \\
        MDocAgent (Qwen3-30B + Qwen2.5-VL-32B)  & 12 & 55.3 & 63.2 & 64.9 & \textbf{84.5} & 66.9  \\
        \midrule
        \multicolumn{7}{c}{\textit{RAG methods (top 10)}} \\
        % \midrule
        M3DocRAG (Qwen2.5-VL-32B) & 10 & 39.7 & 52.2 & 56.7 & 78.6 & 56.8\\
        MDocAgent (Qwen3-30B + Qwen2.5-VL-32B) & 20 & 54.8 & 61.9 & 63.1 & 84.1 & 65.9 \\
        \midrule
        \multicolumn{7}{c}{\textit{Ours (top-10 and top-30)}} \\
        % \midrule
        \textbf{\ours{}} (Qwen3-30B + Qwen2.5-VL-32B)  & 3.2&  59.55&  72.26 &  64.38&  80.31& 69.12 \\
        \textbf{\ours{}} (Qwen3-30B + Qwen2.5-VL-32B) & 3.5&  \textbf{60.58}&  \textbf{72.30}&  \textbf{65.39}&  82.19& \textbf{70.12} \\
        \bottomrule 
    \end{tabular}
    \end{adjustbox}
    \caption{Accuracy(\%) on 4 different DocVQA datasets. We use ColQwen-2.5 as the retrieval model for all methods. \emph{Pg.~Ret.} indicates the actual pages used during generation.}
    \vspace{-0.5em}
    \label{tab:main_results}
\end{table*}

\subsection{Main Results}
\label{sec:mainresult}

\paragraph{Datasets.} 
We evaluate \ours{} on 4 comprehensive PDF document understanding benchmarks, which provide a robust testbed for assessing document understanding at scale across varied document types, lengths, and retrieval complexities:

\textit{1) MMLongBench}~\cite{mmlongbench}: This dataset is designed to test document reasoning over long PDFs, containing complex layouts and multi-modal components. The dataset contains 1073 questions across 135 documents, with an average length of 47.5 pages per document.

\textit{2) LongDocURL}~\cite{longdocurl}: Another large-scale multi-modal benchmark aimed at evaluating document retrieval and reasoning. It has over 33,000 document pages and includes 2,325 question samples.

\textit{3) PaperTab}~\cite{hui2024udabenchmarksuiteretrieval}: It focuses on the extraction and interpretation of the tabular data from the research papers, providing 393 questions from over 307 academic documents.

\textit{4) FetaTab}~\cite{hui2024udabenchmarksuiteretrieval}: A table-based question answering dataset using tables extracted from Wikipedia articles. It presents 1,023 natural language questions across 878 documents, requiring models to generate free-form answers.

\vspace{-0.2em}
\paragraph{Baselines.} We compare with two baselines: (1) \textit{M3DocRAG}~\cite{m3docrag} first uses an image retrieval model to retrieve top-k pages, and then uses a VLM to generate an answer with retrieved pages. (2) \textit{MDocAgent}~\cite{han2025mdocagent} employs both text retrieval model and image retrieval model to retrieve two sets of pages, then top-k pages from both sets will be used for generation.  MDocAgent uses 5 different agents and require both a VLM and a text model. We also include the results of using a VLM to solve the question directly, and results of using VLM with the ground-truth pages included as images (denoted as GT pages), which can be seen as lower and upper bounds.

% \begin{table}[ht!]
%   \centering
%   \caption{All-Match Retrieve Rate on MMLongBench by \ours{} and by embedding retrieval of ColQwen-2.5 (v0.2) on MMLongBench.}
%   \label{tab:all_match}
%   \resizebox{1\linewidth}{!}{%
%   \begin{tabular}{lcc}
%     \toprule
%     \textbf{Retreival method} & \textbf{Avg. Ret. Pages} & \textbf{F1 Score}\\
%     \midrule
%     ColQwen-2.5 & 2  & 38.75\\ 
%     ColQwen-2.5 & 6  & 24.36\\ 
%     ColQwen-2.5 & 10 & 18.38\\ 
%     \midrule
%     % \multicolumn{3}{c}{\textit{Top-10}} \\
%     % \textbf{Ours} (1 iter) & 2.51 & 60.45\\
%     \textbf{Ours} (Top-10)& 3.19 & 61.42\\
%      \textbf{Ours} (Top-30) & 3.46 & \textbf{62.22}\\

%     % \midrule
%     % \multicolumn{3}{c}{\textit{Top-30}} \\
%      % \textbf{Ours} (1 iter) & 2.79 & 62.30\\
%      \bottomrule
%   \end{tabular}%
%   }
% \end{table}

\begin{table}[ht!]
  \centering
  \caption{All-Match Retrieve Rate, and Page-level F-1 Score on MMLongBench (See Section~\ref{sec:metrics} for calculation). We present the results for ColQwen (used by M3DocRAG and MDocAgent) and our retrieval.}
  \label{tab:all_match}
  \resizebox{1\linewidth}{!}{%
  \begin{tabular}{lccc}
    \toprule
    \textbf{Method} & \textbf{Avg Ret. Pages}& \textbf{All Hit \%} & \textbf{F1 Score}\\
    \midrule
    ColQwen-2.5 & 2  & 64.12& 38.75\\ 
    ColQwen-2.5 & 6  & 76.42& 24.36\\ 
    ColQwen-2.5 & 10 & \textbf{83.60}& 18.38\\
    \textbf{Ours} (top-10)& 3.19 &  65.72&61.42\\
     \textbf{Ours} (top-30) & 3.46 & 67.37 & \textbf{62.22}\\
    \bottomrule
  \end{tabular}%
  }
      \vspace{-0.5em}
\end{table}

\vspace{-0.3em}
\paragraph{Metrics.} For this experiment, we evaluate model performance with \textit{Binary Correctness (Accuracy)}. We classify each model response as either correct or incorrect and compute the accuracy as the ratio of correct responses to the total number of questions. We use \textbf{GPT-4.1} as an automatic evaluator to judge response correctness against ground truth answers and set the temperature to 0.
% ~\todo{@Chelsi: We use binary correctness and calculate the accuracy. For the second table, since MMLongBench has unanswerable questions, we can also calculate a F1-score to understand the hallucination of LLMs}
% We evaluate model performance using the following tow metrics:
% \begin{itemize}
% \item \textbf{Binary Correctness (Accuracy):} 

%     \item 
% \end{itemize}

% (ColBERTv2~\cite{colbert}) (\texttt{Qwen2.5-VL-32B-Instruct}) 
% \todo{@Chelsi, explain MDocAgent, M3DocAgent, please be clear on 1. how they retrieve 2. how they generate answer.} \yr{consider add PlanRAG+MDocAgent and Chain-of-Note+MDocAgent}
\vspace{-0.3em}
\paragraph{Implementation Details.} We use the same models for \ours{} and baselines for rigorous comparison. For visual embedding model, we use \texttt{ColQwen-2.5} for all methods, which is the latest model trained with CoPali~\cite{colpali}'s strategy (See Table~\ref{tab:all_match} for a comparison with CoPali), and we use \texttt{Qwen2.5-VL-32B-Ins} whenever a VLM is needed.
For MDocAgent, we use ColBERTv2~\cite{colbert} as the text retrieval model following the original paper, and \texttt{Qwen3-30B-A3B} as the text model. 
For~\ours{}, we use \texttt{Qwen2.5-VL-32B-Ins} for per-page summarization during pre-processing. Note that the summarization only needs to be performed once. We use \texttt{Qwen3-30B-A3B} to for page retrieval. For baselines, we test with top-k set to 2, 6, 10. For our method, we set top-k to 10 and 30 for embedding retrieval. All prompts used in our method is shown in Appendix~\ref{sec:prompt}.

% \textbf{1. MDocAgent.} We use Colbert for text retrieval (as in original paper). For image retrieval, we test both ColQwen-2.5 (latest) and ColPali (used in original paper). For generation, we use \texttt{Qwen2.5-32B} for all agents that require VLM, and use Llama-3-8B for text agent \yr{Consider use Llama-4-Scout if our method is good}. \textbf{2. M3DocRAG} We use \texttt{ColPali} for text retrieval and \texttt{Qwen2.5-32B} for generation.
% \yr{If our final method doesn't have retrieval, we can fix image retrieval to ColPali.} \textbf{3. Our Method.}

% \todo{write analysis based on the results}
% Table 1 presents accuracy results across four DocVQA benchmarks under various methods, including pure LVLMs, multi-modal retrieval-augmented generation (RAG) baselines (M3DocRAG \cite{m3docrag} and MDocAgent \cite{han2025mdocagent}), and the proposed ~\ours{} framework. 
% For MDocAgent, we use \texttt{Qwen2.5-32B} as the
\vspace{-0.3em}
\paragraph{Results Analysis} Table~\ref{tab:main_results} shows that ~\ours{} achieves the highest average accuracy of 70.12\%, outperforming all the baselines with different top-k retrieval settings. On MMLongBench and LongDocURL, which contain long, diverse, and multi-modal documents, our proposed method significantly outperforms MDocAgent by +5.3\% and +9.1\%, respectively. These gains highlight strength in addressing complex queries that require aggregating information dispersed across different sections of a document. However, on FetaTab, a heavily table-centric dataset,~\ours{} performs lower than MDocAgent. We attribute this to MDocAgent's explicit multi-agent design, which uses a dedicated image agent to focus on another modality (table grids) and is especially effective for this specific type of table-based QA. In contrast, \ours{} treats pages as images to feed into one single reasoner agent. Thus, \ours{} is more robust and effective across questions that require diverse evidence types.

Table~\ref{tab:main_results} also lists the average number of pages each system retrieves. \ours{} needs only 3.5 pages per question yet achieves the best overall accuracy. By contrast, MDocAgent attains 59.6\% accuracy when it reads 4 pages, which is about 10 percentage points below our method. Notably, both MDocAgent and M3DocRAG reach their peak accuracy at top‑k=6 rather than 10, implying that indiscriminately adding pages can hurt performance. To understand this effect, Table~\ref{tab:all_match} reports two retrieval metrics. 1) The all‑hit rate gauges coverage, the fraction of questions for which the entire gold evidence set appears among the retrieved pages. 2) The page‑level F1 score captures efficiency, rewarding systems that surface the right pages while avoiding noise. For \texttt{ColQwen‑2.5}, raising k from 2 to 10 boosts coverage but reduces F1, showing that many of the extra pages are irrelevant. Thus, top-k=6 reflects a better balance between coverage and conciseness, which in turn yields higher answer accuracy for the agent baselines. In contrast, \ours{} attains nearly the same coverage as \texttt{ColQwen‑2.5} at k=2 yet more than doubles its F1, demonstrating that our retriever supplies almost all necessary evidence with far less clutter. 
Overall, \ours{} delivers the best coverage‑versus‑conciseness trade‑off while avoiding trial-and-error to find the best top-k retrieval numbers, giving the reasoner everything it needs while keeping the reading budget minimal. 

% . Although our method may be less effective at this specific dataset, but it outperforms other 
% images and summaries in a unified loop and relies on summary-based refinement
% Adding more table-aware processing could further improve ~\ours{}'s performance on tabular datasets.
 % This targeted reasoning allows it to more precisely capture fine-grained spatial relationships.
% This performance demonstrates the effectiveness of ~\ours{} iterative retrieval-refinement mechanism that integrates both embedding-based and summary-based evidence.
% Through iterative refinement, the proposed method incrementally gathers more relevant and contextually aligned information. Notably, its performance approaches that of an oracle setup in which the ground truth evidence pages are provided in advance, indicating that ~\ours{}'s retrieval mechanism is capable of approximating high-quality evidence selection.

% Overall, ~\ours{} demonstrates strong generalization and surpasses the existing RAG and multi-agent systems in average accuracy, while maintaining a simpler architecture and leveraging a unified reasoning framework and using a modern vision-language model.

\begin{figure*}[ht!]
    \centering
    \includegraphics[width=0.9\textwidth]{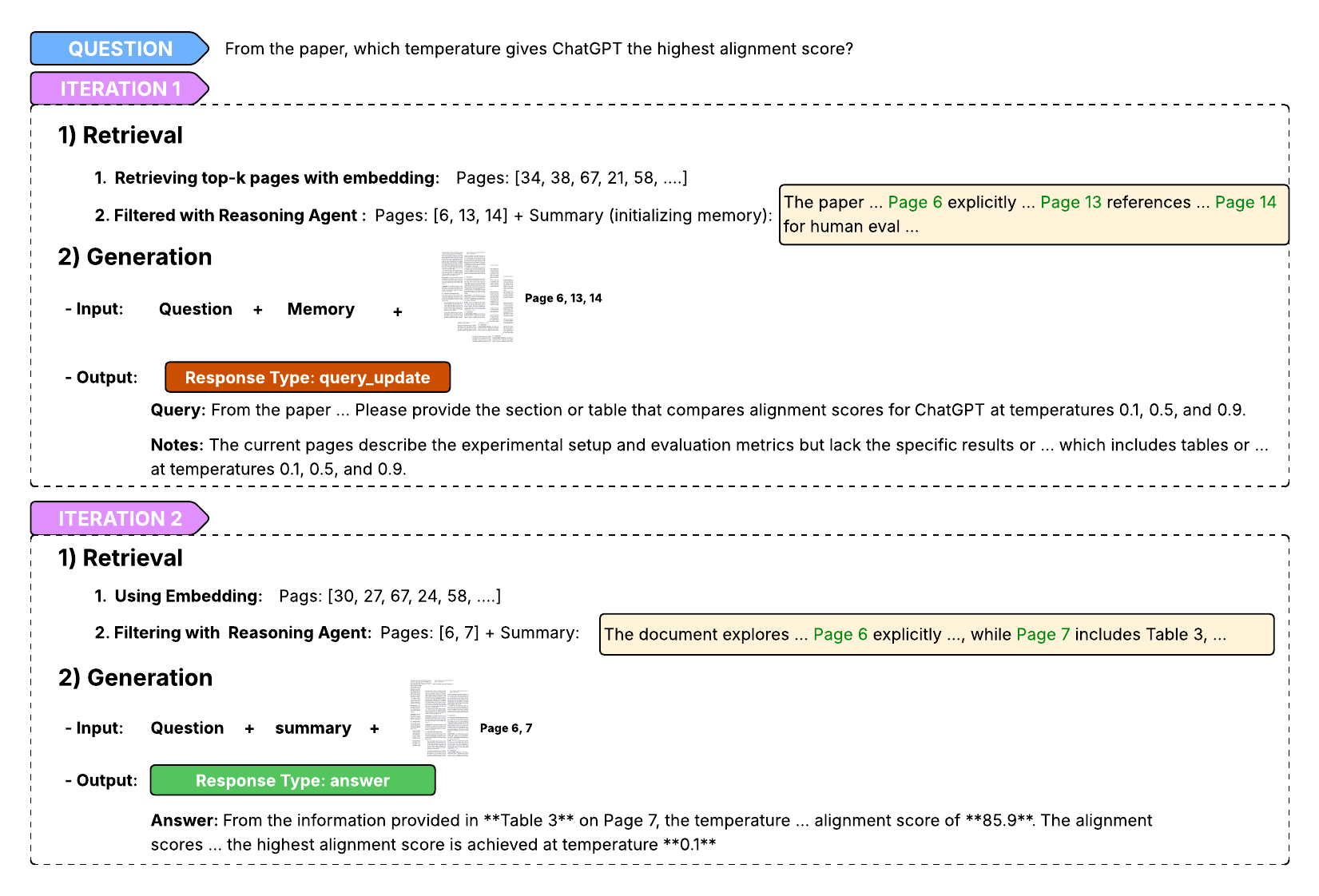}
    \caption{An example run of \ours{}'s iterative reasoning solving a question. In the first round, the agent retrieves Pages 6, 13, and 14 based on embedding and summary-based filtering. However, the retrieved pages only describe the experimental setup and evaluation metrics without giving exact alignment scores. The agent identifies this gap and generates a refined query asking specifically for a section or table comparing scores at different temperatures. This updated query retrieves Page 7, which contains Table 3 with the required information, allowing the agent to correctly answer that temperature 0.1 yields the highest alignment score (85.9).}
    \label{fig:example}
\end{figure*}

\begin{table*}[t!]
\centering
\small
\begin{tabular}{lccccccccc}
\toprule
\multirow{2}{*}{Method} &
\multicolumn{5}{c}{Evidence Source} &
\multicolumn{3}{c}{Evidence Page} &
\multirow{2}{*}{ACC} \\
\cmidrule(lr){2-6} \cmidrule(lr){7-9}
 & TXT & LAY & CHA & TAB & FIG & SIN & MUL & UNA &  \\
\midrule
\multicolumn{10}{c}{\textit{Qwen2.5-VL-7B-Instruct + Qwen-3-8B}}\\
\midrule
VLM + \textcolor{blue}{GT pages} & 51.32 & 45.38 & 37.71 & 40.09 & 47.83 & 58.90 & 35.01 & 77.97 & 54.99 \\
M3DocRAG (top-6)                & 43.21 & 39.98 & 36.05 & 31.60 & 42.01 & 55.46 & 24.78 &  8.72 & 35.50 \\
MDocAgent (top-6)               & 47.04 & 38.98 & \textbf{47.09} & \textbf{41.04} & 39.93 & \textbf{59.45} & 28.57 & 33.49 & 43.80 \\
\textbf{Ours}                   & \textbf{49.67} & \textbf{42.02} & 44.57 & 37.79 & \textbf{42.14} & 58.69 & \textbf{31.65} & \textbf{62.11} & \textbf{50.42} \\
\midrule
\multicolumn{10}{c}{\textit{Qwen2.5-VL-32B-Instruct + Qwen-3-30B-A3B}}\\
\midrule
VLM + \textcolor{blue}{GT pages} & 63.25 & 66.39 & 58.86 & 65.44 & 57.53 & 72.60 & 55.46 & 77.53 & 67.94 \\
M3DocRAG                         & 46.69 & 41.53 & 45.35 & 39.15 & 43.75 & 58.61 & 30.61 & 22.48 & 41.80 \\
MDocAgent                        & 57.49 & 50.00 & 54.65 & \textbf{56.13} & \textbf{52.78} & 68.70 & \textbf{42.86} & 45.41 & 55.30 \\
Chain-of-Notes$^\dagger$         & 36.75 & 35.29 & 38.46 & 32.26 & 33.44 & 49.59 & 21.69 & 50.00 & 40.45 \\
Plan$^*$RAG$^\dagger$            & 46.03 & 36.13 & 43.75 & 38.71 & 37.12 & 54.88 & 25.35 & 23.89 & 38.58 \\
\textbf{Ours}                   & \textbf{59.93} & \textbf{51.26} & \textbf{54.86} & 51.15 & 51.17 & \textbf{70.76} & 39.22 & \textbf{67.40} & \textbf{59.55} \\
\bottomrule
\end{tabular}
\caption{Performance with different models on \textbf{MMLongBench}. We present detailed accuracy for questions with five different evidence sources: text (TXT), layout (LAY), chart (CHA), table (TAB), and figure (FIG); different numbers of evidence pages (single (SIN), multiple (MUL), unanswerable (UNA), and average accuracy.
We also test two RAG methods originally proposed for knowledge QAs on MMLongBench (labeled with $^\dagger$). }
\vspace{-0.5em}
% \todo{@Yiran: what is $^\dagger$ for? footnote? Shall we highlight numbers? }
\label{table:mmlongbench}
\end{table*}

\paragraph{Qualitative Analysis} 
% As shown in Figure \ref{fig:example}, initially, the agent identified potentially relevant sections of the document concerning ChatGPT's temperature and alignment scores (Pages 6, 13, 14). However, its analysis during the first iteration revealed that while the general context was correct, the precise data needed—the specific temperature yielding the highest alignment score—was not explicitly available in the retrieved content. Rather than concluding prematurely, the agent exhibited a key strength: it diagnosed the information gap and formulated a more targeted follow-up. This led to the generation of an updated query specifying the need for "the section or table that compares alignment scores for ChatGPT at temperatures 0.1, 0.5, and 0.9." This refinement proved crucial. In the subsequent iteration, this more precise query directed the agent to Page 7, which contained Table 3. The agent then successfully processed this tabular data, pinpointing that a temperature of 0.1 produced the highest alignment score (85.9).

As illustrated in Figure\ref{fig:example}, ~\ours{} demonstrates its ability to reason iteratively. Initially, it retrieves pages that are broadly relevant but lacking specific details needed to answer the question. Recognizing the gap, the agent refines the query to target missing information, retrieves the precise page containing the relevant table, and successfully answers the question. This example highlights how ~\ours{} detects incomplete evidence and adaptively improves retrieval to resolve complex queries.

% \todo{@Yifan: We provide one example run using our method, present the summarizes, retrieval pages, and the regenerated query from iteration.}
\vspace{-0.3em}
\subsection{Results with different models}
\vspace{-0.3em}

\label{sec:diff_model}

In Table~\ref{table:mmlongbench}, we test with smaller models (\texttt{Qwen2.5-VL-7B-Instruct} + \texttt{Qwen-3-8B}) with detailed results on MMLongBench to further validate our method. Note that \texttt{Qwen-3-8B} are text-only models and used in MDocAgent (Text Agent) and our method (for retrieval). Our method outperforms all baselines in terms of avg. accuracy (ACC) for both models. Under the smaller 7B/8B model setting, our method achieves 50\% overall accuracy, improving over MDocAgent by +6.62 points, which is a bigger gap compared to using larger models (+4.15 points). When broken down by evidence source, our model achieves the best performance on three out of five modalities. We note that MDocAgent are competitive on charts and tables with specialized agents, which is consistent with our observation and analysis in Section~\ref{sec:mainresult}. When broken down by number of evident pages, our methods have similar results compared with MDocAgent on multi-page (MUL) and single-page (SIN) reasoning with different models. However, \ours{} achieves much better results on unanswerable questions, which is used to test hallucinations, showcasing its ability to abstain from guessing when no valid evidence is present.

\vspace{-0.3em}
\subsection{Other RAG methods}
\vspace{-0.3em}
\label{sec:other_rag}
We also adopt and evaluate two RAG methods that originally focus on knowledge question answering tasks: \textbf{(1) Plan$^*$RAG} \cite{planrag}: first decomposes a question into sub-queries that form a directional acyclic graph (DAG). It starts with solving the leaf sub-queries, and incorporates the previous subquery+answer when solving the next queries, until the original question. This features the query-decomposition and augmented process strategies, which are common in RAG methods.
\textbf{(2) Chain-of-Notes}~\citep{chainofnote} taking notes of retrieved paragraphs and then using them for more precise generation. We do the following to adapt them to our setting: we use ColQwen2.5 to retrieve document pages, and use VLM for generation, which is the same as other baselines.

Table~\ref{table:mmlongbench} reports the performance of the two RAG baselines when paired with \texttt{Qwen2.5‑VL‑32B}. Both Chain‑of‑Note and Plan$^\ast$RAG lag behind approaches designed specifically for DocVQA, indicating that simply transplanting text‑oriented knowledge-based RAG techniques is insufficient for this domain. From our analyses, we also observe potential failure reasons for each method: (1) Since Chain-of-Note uses page-level image summary, it can miss finer details like exact numbers in tables or exact words in charts and layouts. Also, one summary per page can be too general, making it hard to reason across multiple pages or give precise answers, yielding only 40.4\% accuracy. (2) Plan$^*$RAG uses full-page images and breaks the main question into sub-questions using a query decomposition step. However, the acyclic graph it builds is often not accurate, leading to off-target sub-queries. For each one, it retrieves top-k image pages, generates answers, and then summarizes them. This multi-step pipeline adds complexity and increases error propagation. 

\vspace{-0.3em}
\subsection{Additional Analysis of \ours{}}
\label{sec:additional_result}

In this section, we do more experiments to decompose and analyze our method. 

\begin{table}[ht!]
  \centering
  \begin{tabular}{ccc}
    \toprule
    \textbf{Top-k} & \textbf{Avg. Page Used} & \textbf{Acc.} \\
    \midrule
    2  & 2.15& 56.66\\
    6  & 2.75& 58.25\\
    10 & 3.19& 59.55\\
    30 & 3.46& 60.58\\
    \bottomrule
  \end{tabular}
  \caption{Our method with different top-k numbers for embedding retrieval on MMLongBench. \textit{Avg. Page Used} denotes the actual number of pages seen by the reasoner agent.}
  \vspace{-0.4em}
  \label{tab:ablation_pages_accuracy}
\end{table}

\textbf{Varying top-k for embedding retrieval.} In \ours{}, we first retrieve top-k pages based on embeddings, and then use a LLM to re-rank them based on summaries. With retrieval, we can filter and bound the maximum number of pages before re-ranking. In this experiment, we test our method with different numbers of top-k pages retrieved through embedding. 
The increase in Top-k gives the LLM retrieval agent more space to select the most closely related pages that were not correctly identified by the embedding-based retrieval method.
We didn't see the retrieval agent select significantly more pages in the setting where K is large.
This means the agent is dynamically deciding which pages are truly relevant to the given query.
% \todo{@yifan: fill Table~\ref{tab:ablation_pages_accuracy}, explain the result in few sentences}

\textbf{Results with different iterations.} 
Table~\ref{tab:iteration} illustrates the benefits of our iterative refinement strategy on MMLongBench. The observed trend shows that additional iterations allow \ours{} to progressively enhance understanding and locate crucial information initially missed. This targeted re-querying leads to improved accuracy, while the decreasing number of query updates indicates the system is either satisfying the information need or recognizing when an answer cannot be found within the document.

\begin{table}[]
\centering
\begin{tabular}{c|ccc}
\toprule
 Iteration&  1 & 2&3\\
\midrule
 Accuracy&  58.62& 59.27&59.55 \\
\# Query Upate &  182& 121&97\\
\bottomrule
 \end{tabular}
\caption{Performance of \ours{} on MMLongBench across different iterations, showing accuracy and number of query updates.}
    \vspace{-0.6em}
\label{tab:iteration}
\end{table}

\vspace{-0.6em}
\section{Conclusion}
\vspace{-0.3em}

We present \ours{}, an effective framework for multi‑modal document QA. \ours{} consists of an efficient retrieve module that utilizes both dense‑vector embedding and summary, to retrieve the pages efficiently, and a reasoning agent that can detect and remedy missing evidence iteratively. Empirical results across 4 DocVQA benchmarks confirm that \ours{} surpasses prior RAG‑style systems and multi‑agent baselines with fewer components and fewer page retrievals. These results highlight how modern VLMs can be used on retrieval‑augmented multi-modal reasoning.

% can be used to reasoning over long, multi‑page, multi‑modal documents, pointing toward future work on structured reasoning and open‑domain settings.

\section{Limitations} In this work, we only experiment with single-document VQAs, while the embedding retrieval method can be readily extensible to retrieve from the whole document database. We believe there are still many interesting research questions under this scenario. We focus on test-time scaling methods instead of training, and we think more RAG methods that require training~\cite{selfrag, rqrag} can be utilized for this task. Finally, graph-based database and retrieval methods are also future directions to explore\cite{edge2024local, liu2025hm}.

% Graph Databases. Training. Multi-Document VQA

% 1.
% 2. Retrieved top k + Answering
%     2. 1 (MDoc)
% 3. Retrieved top k + Reranking + Answering
% 4. Retrieved top k + Reranking + Answering
%     - > If need more pages, generate subqueries and retrieve

% \bibliographystyle{numeric}
% \clearpage
\bibliographystyle{acl_natbib}

\clearpage
\newpage
\appendix

\section{Appendix}
\label{sec:appendix}

\begin{table}[ht!]
  \centering
  \caption{All-Match Retrieve Rate on MMLongBench with two retrieve models. A question is all-match if all ground-truth evident pages is present in the retrieved pages. Note that ColQwen-2.5 (v0.2) is trained with strategy introduce by CoPali.}
  \label{tab:all_match}
  \begin{tabular}{lcc}
    \toprule
    \textbf{Model} & \textbf{Top-K} & \textbf{Match Rate \%} \\
    \midrule
    ColQwen-2.5 & 2  & 54.55\\ 
    CoPali      & 2  & 28.74 \\ 
    ColQwen-2.5 & 6  & 70.13 \\ 
    CoPali      & 6  & 44.35 \\ 
    ColQwen-2.5 & 10 & 79.22 \\ 
    CoPali      & 10 & 55.15 \\ 
    \bottomrule
  \end{tabular}
\end{table}

% \begin{figure*}
%     \centering
%     \small
% \begin{llmprompt}
% Your prompt
% \end{llmprompt}
%     \caption{Additional prompt added when testing with \texttt{o3-mini}, \texttt{o4-mini}, \texttt{o1-mini}.}
%   \vspace{-1em}
%     \label{fig:additional_prompt}
% \end{figure*}

\subsection{Pilot Study}
\label{sec:pilot}

We perform a pilot experiment on MMLongBench to understand how VLM performs on DocVQA problems. To compare, we test \texttt{Qwen2.5-VL-32B} with no evidence page and with ground-truth evidence pages provided by the dataset. To understand how different modalities of evidences affect the results, we also input image of the pages, text of the pages (extracted with PDF tools), and both text and image of the pages. We find that using the image form of ground-truth pages is crucial, since there is 25\% accuracy gap between image-based and text-based input. Combining the two forms can further boost the performance, but are not significant.

% how can VLMs answer the questions, with different modalities of input?

% \paragraph{Key Conclusion.} If we can find the evident pages correctly, recent VLM can answer the questions well.

\begin{table}[ht!]
  \centering
  \caption{Model accuracies by input type (values to be filled)}
  \label{tab:model_accuracy}
  \begin{tabular}{llc}
    \toprule
    \textbf{Doc Type} & \textbf{Model} & \textbf{Accuracy (\%)} \\
    \midrule
    N/A & Qwen2.5‑VL-32B     & 22.18\\
    % Random Both & Qwen2.5‑VL-32B   & -- \\
    GT Image & Qwen2.5‑VL-32B     & 67.94\\
    % GT Image & Gemma-3‑32B & -- \\
    % Image & Gemma3‑12B & -- \\
    % GT Image & Qwen2.5‑VL-32B        & -- \\
    % GT Text  & Gemma-3‑32B & -- \\
    % Text  & Gemma-3‑12B & -- \\
    GT Text  & Qwen2.5‑VL-32B        & 42.40\\
    % GT Both  & Gemma-3‑32B & -- \\
    % Both  & Gemma-3‑12B & -- \\
    GT Both  & Qwen2.5‑VL-32B        &  69.06
\\
    \bottomrule
  \end{tabular}
\end{table}

\subsection{Usage of AI assistant}
We use AI assistant to help debug code and build utility functions. We also use AI assistant to refine writing.

\subsection{Detailed Retrieval Metric Calculation}
\label{sec:metrics}

Let $\mathcal{Q}$ be the set of $N$ evaluation questions.  
For every question $q\in\mathcal{Q}$ we denote by
\[
  G_q \subseteq \mathcal{D}, 
  \qquad 
  R_q \subseteq \mathcal{D}
\]
the {\em gold} set of truly relevant pages and the {\em retrieved} set
(the top–$k$ pages produced by the system).

%--------------------------------------------------
\paragraph{All‑hit Rate (Coverage)}
The {\em all‑hit rate} measures the proportion of questions for which
{\em every} gold page is retrieved:
\[
  \mathrm{AllHit} \;=\;
  \frac{\bigl|\{\,q\in\mathcal{Q}\;:\;G_q\subseteq R_q\,\}\bigr|}{N}.
\]
Because a single missing page makes a query count as a failure,
All Hit captures strict evidence coverage.

%--------------------------------------------------
\paragraph{Page‑level F\textsubscript{1} (Retrieval Efficiency)}
Retrieval may also be viewed as a binary decision for each candidate
page (gold vs.\ non‑gold).  
For every question we compute \emph{precision} and \emph{recall},
abbreviated $P_q$ and $R_q$:
\[
  P_q \;=\; \frac{|G_q \cap R_q|}{|R_q|},
  \qquad
  R_q \;=\; \frac{|G_q \cap R_q|}{|G_q|}.
\]
Their harmonic mean gives the question‑level F\textsubscript{1}:
\[
  \mathrm{F1}_q \;=\;
  \begin{cases}
    \dfrac{2\,P_q\,R_q}{P_q + R_q}, & \text{if } P_q + R_q > 0,\\[6pt]
    0, & \text{otherwise.}
  \end{cases}
\]
Macro‑averaging over questions yields the final score:
\[
  \mathrm{PageF1} \;=\; \frac{1}{N}\sum_{q\in\mathcal{Q}} \mathrm{F1}_q.
\]

% \paragraph{Interpretation.}
% \emph{All‑hit rate} answers “Did we surface {\em all} necessary evidence?”
% while \emph{page‑level F\textsubscript{1}} balances conciseness and
% completeness—rewarding systems that return exactly the relevant pages,
% no more and no fewer.  Reporting both scores gives a fuller picture of
% retrieval behavior.

\subsection{Prompts Used in \ours{}}
\label{sec:prompt}
In Figure~\ref{fig:page_index_prompt}, we show the prompt for pre-processing each page. In Figure~\ref{fig:retrieval_prompt}, we show the prompt to retrieve pages based on reasoning. In Figure~\ref{fig:qa_prompt}, we should the prompt for the reasoner agent.

\begin{figure*}[t]
    \centering
    % \small
\begin{llmprompt}
\textbf{Page Index Prompt:}

You are tasked with creating a comprehensive summary of a given page from a document. Your summary should focus on extracting and describing the main content, tables, figures, and images present on the page.
\\

Raw text extracted from the retrieved pages (without visual information): \\
\verb|<page_text>| 
\\
\{PAGE\_TEXT\} 
\\
\verb|</page_text>|

Please follow these steps to create your summary:

1. Carefully read and analyze the page content.\\
2. Identify the main topics, key points, and important details presented on the page.\\
3. Note any tables, figures, charts, diagrams, or images on the page and briefly describe their content and purpose.\\
4. Create a structured summary that captures:\\
   - The essential textual information from the page \\
   - Descriptions of any visual elements (tables, figures, images, etc.) \\
   - Any particularly notable or unique information\\

Present your summary within <summary> tags. The summary should be concise yet comprehensive, typically 5-8 sentences for text-only pages, with additional sentences as needed to describe visual elements.

For visual elements, please use these specific tags:\\
- \verb|<table_summary>| for descriptions of tables\\
- \verb|<figure_summary>| for descriptions of figures, charts, graphs, or diagrams\\
- \verb|<image_summary>| for descriptions of photos, illustrations, or other images\\

Example structure:\\
\verb|<summary>|
[Main text content summary here]\\

\verb|<table_summary>| Table 1: [Brief description of what the table shows] \verb|</table_summary>|

\verb|<figure_summary>| Figure 2: [Brief description of what the figure depicts] \verb|</figure_summary>|

\verb|<image_summary>| [Brief description of image content] \verb|</image_summary>|
\verb|</summary>|
\end{llmprompt}
    \caption{Page indexing prompt used to extract structured information from document pages.}
    \label{fig:page_index_prompt}
\end{figure*}

\begin{figure*}[t]
    \centering
    % \small
\begin{llmprompt}
\textbf{Page Retrieval Prompt:}

You are a document understanding agent tasked with identifying the most promising page(s) for a given user query. You will be presented with summaries of each page in a document and a user query. Your task is to determine which page(s) should be examined in detail in a subsequent step. \\

First, review the summaries of each page in the document: \\

\verb|<page_summaries>|
{PAGE\_SUMMARIES}
\verb|</page_summaries>|

Now, consider the following user query: \\

\verb|<user_query>|

{USER\_QUERY}

\verb|</user_query>|

Important context about your task:\\
1. You are performing an initial screening of pages based on limited information (summaries only).\\
2. The pages you select will be analyzed in depth by another agent who will have access to the full page content.\\
3. These summaries are inherently incomplete and may miss details that could be relevant to the query.\\
4. It's better to include a potentially relevant page than to exclude it at this stage.\\

To determine which pages warrant closer examination:\\

1. Identify keywords, topics, and themes in the query that might appear in the document.\\
2. Select any page(s) whose summaries suggest they might contain information related to the query.\\
3. Be inclusive rather than exclusive - if a page seems even somewhat related or contains terminology connected to the query, include it for further analysis.\\
4. Always select at least one page, even if the connection seems tenuous - the detailed examination will determine true relevance.\\
5. The page order should be from most relevant to less relevant in your answer.\\

Additionally, create a comprehensive document-level summary that addresses the user query based on your understanding of the entire document. This summary should:\\
1. Provide a high-level perspective on how the document relates to the query\\
2. Synthesize relevant information across multiple pages\\
3. Highlight key concepts, definitions, or facts from the document that pertain to the query\\
4. Outline a strategic approach to solving the query based on the document's content\\
5. Identify potential solution paths and the types of information that should be prioritized\\
6. Do not be too certain about the conclusions drawn from the summaries, as they may not capture all relevant details\\
7. Be concise but informative (5-8 sentences)\\

After your analysis, provide your final answer in the following format:\\

\verb|<document_summary>|
[A comprehensive summary addressing how the document relates to the user query...]
\verb|</document_summary>|

\verb|<selected_pages>|
[List the indices of selected pages, separated by commas if there are multiple] \\
\verb|</selected_pages>|
\end{llmprompt}
    \caption{Prompt for selecting top pages to retrieve for downstream reasoning.}
    \label{fig:retrieval_prompt}
\end{figure*}

\begin{figure*}[t]
    \centering
    \small
\begin{llmprompt}
\textbf{Question Answering Prompt:}

You are an AI assistant capable of analyzing documents and extracting relevant information to answer questions. You will be provided with document pages and a question about these pages.\\

Consider this question about the document:\\
\verb|<question>|
{QUESTION}
\verb|</question>|

Document level summary:\\
\verb|<document_summary>|
{DOCUMENT\_SUMMARY}
\verb|/document_summary>|

The page numbers of the CURRENT RETRIEVED PAGES that you should analyze:\\
\verb|<retrieved_pages>|
{RETRIEVED\_PAGE\_NUMBERS}
\verb|</retrieved_pages>|

Raw text extracted from the retrieved pages (without visual information):
\verb|<page_text>|
{PAGE\_TEXT}
\verb|</page_text>|

IMPORTANT: Images of the retrieved pages are attached at the end of this prompt. The raw text extracted from these images is provided in the \verb|<page_text>| tag above. You must analyze BOTH the visual images AND the extracted text, along with the 
\verb|<document_summary>|, to fully understand the document and answer the question accurately.\\

\verb|<scratchpad>|
1. List key elements from text and images \\
2. Identify specific details that relate to the question \\
3. Make connections between the document information (from both images, text, summary) and the question
4. Determine if the provided information is sufficient to answer the question
5. If you believe other pages might contain the answer, be specific about which content you're looking for that hasn't already been retrieved
\verb|</scratchpad>|

CRITICAL INSTRUCTION: First carefully check if:\\

The pages listed in \verb|<retrieved_pages>| are already the specific pages that would contain the answer to the question\\
The specific tables, figures, charts, or other elements referenced in the question are already visible in the current images\\
The document summary explicitly mentions the content you're looking for\\
Do not request these same pages or elements again in a query update.\\

Based on your analysis in the scratchpad, respond in one of three ways:\\

If the provided pages contain sufficient information to answer the question, or if the document summary clearly indicates the answer to the question is that something does not exist:\\
\verb|<answer>|
Your clear and concise response that directly addresses the question, including an explanation of how you arrived at this conclusion using information from the document.
\verb|</answer>|

If based on the document summary and current pages, you're confident the entire document likely doesn't contain the answer, OR if the specific pages/tables/figures/elements that should contain the answer are already in the current context but don't actually contain relevant information:\\
\verb|<not_answerable>|
The document does not contain the information needed to answer this question.
\verb|</not_answerable>|

If based on the document summary, you believe the answer exists in other parts of the document that haven't been retrieved yet:\\
\verb|<query_update>|
[Provide a rewritten long query that PRESERVES THE ORIGINAL MEANING of the question but adds specific details or keywords to help retrieve new relevant pages. The information retrieved from this new query must directly answer the original question.]
\verb|</query_update>|

\verb|<notes>|
[IF using query\_update, provide concise notes about what you've learned so far, what information is still missing, and your reasoning for the updated query. These notes will be appended to the document summary in the next iteration to maintain context across searches.]
\verb|</notes>|

Usage guidelines:\\

Use <answer> when you can answer the question with the provided pages, OR when you can determine from the document summary that the answer is that something doesn't exist.\\

Use <not\_answerable> when either:
The document summary and current pages together suggest the document as a whole doesn't contain the answer\\
OR the specific pages that should logically contain the answer are already provided in <retrieved\_pages> but don't actually have the relevant information\\

OR specific tables, figures, charts, or elements mentioned in the question are visible in the current pages but don't contain the information being asked for
\\
Use <query\_update> ONLY when seeking information you believe exists in other pages that have NOT already been retrieved. Never request pages that are already listed in <retrieved\_pages> or elements already visible in the current context.
When creating a <query\_update>, you MUST preserve the original meaning and intent of the question while adding specific details, keywords, or alternative phrasings that might help retrieve the necessary information. The answer to your new query must directly answer the original question.
When using <query\_update>, ALWAYS include the <notes> tag to summarize what you've learned so far and explain your reasoning for the updated query.

Your response must include both the <scratchpad> tag and exactly one of the following tags: <answer>, <not\_answerable>, or <query\_update>. If you use <query\_update>, you must also include the <notes> tag.

\verb|<answer>| / \verb|<not_answerable>| / \verb|<query_update>|
\end{llmprompt}
    \caption{Prompt used during the question-answering stage, leveraging both extracted text and page images.}
    \label{fig:qa_prompt}
\end{figure*}

\end{document}